\documentclass[letterpaper, 10 pt, conference]{ieeeconf}
\IEEEoverridecommandlockouts                              

\usepackage[OT1]{fontenc}
\usepackage{cite}
\usepackage{amsmath,amssymb,amsfonts}
\usepackage{algorithmic}
\usepackage{graphicx}
\usepackage{textcomp}
\usepackage{xcolor}
\usepackage{verbatim}
\usepackage{bm}
\usepackage{amsmath}
\usepackage{mathtools}
\usepackage{commath}
\usepackage{amsmath}
\usepackage{mathtools}
\usepackage{afterpage}
\usepackage{amsmath}
\usepackage{mathtools}
\usepackage{commath}
\usepackage{amsmath,amssymb,amsfonts}
\newcommand{\me}{\mathrm{e}}

\usepackage{booktabs}
\usepackage{makecell}

\usepackage{subfig}

\long\def\/*#1*/{}

\def\BibTeX{{\rm B\kern-.05em{\sc i\kern-.025em b}\kern-.08em
    T\kern-.1667em\lower.7ex\hbox{E}\kern-.125emX}}
\begin{document}

\title{\LARGE \bf Making Sense of Touch: Unsupervised Shapelet Learning in Bag-of-words Sense\\
\/*{\footnotesize \textsuperscript{*}Note: Sub-titles are not captured in Xplore and
should not be used}
\thanks{Identify applicable funding agency here. If none, delete this.}*/
}

\author{Zhicong Xian$^{1}$, Tabish Chaudhary$^{2}$ and J\"urgen Bock$^{3}$
\thanks{*This work was supported by KUKA Germany GmbH}
\thanks{Zhicong Xian, J\"urgen Bock and Tabish Chaudhary are with Corporate Research, KUKA Germany GmbH,
Zugspitzstrasse 140, 86165 Augsburg, 
{\tt\small \{zhicong.xian@kuka.com}}
}
\/*
\thanks{$^{1}$Zhicong Xian is with Corporate Research, KUKA Germany GmbH,
        Zugspitzstrasse 140, 86165 Augsburg, 
        {\tt\small zhicong.xian@kuka.com}}%
\thanks{$^{2}$Tabish Chaudhary is student with with Corporate Research, KUKA Germany GmbH,
        Zugspitzstrasse 140, 86165 Augsburg}%
\thanks{$^{3}$J\"urgen Bock is a cluster leader with with Corporate Research, KUKA Germany GmbH,
        Zugspitzstrasse 140, 86165 Augsburg,
        {\tt\small juergen.bock@kuka.com}}%
*/

\maketitle
\thispagestyle{scrheadings} 
\ieeefootline{Workshop on New Advances in Brain-Inspired Perception, Interaction and Learning,\\
International Conference on Robotics and Automation 2020}
\begin{abstract}
A robot manipulation task often requires the correct perception of an interactive environment. Environment perception relies on either visual or haptic feedback. During a contact rich manipulation, vision is often occluded, while touch sensing proves to be more reliable. And a force sensor reading often entails abundant time series segments that reflect different manipulation events. These discriminating time series sub-sequences are also referred to as shapelets. The discovery of these shapelets can be considered as a clustering problem and the distance of sample time series to these shapelets is essentially a feature learning problem. Additionally, shapelets can also be considered as dictionaries in compressed sensing. This paper proposes a neural network with t-distributed stochastic neighborhood embedding as a hidden layer (NN-STNE) to project a long time series into a membership probability to a set of shorter time-series sub-sequences, i.e., shapelets. In this way, the dimensions of input data can be reduced. To preserve the local structure within data in the projected lower-dimensional space as in its original high dimensional space, a Gaussian kernel-based mean square error is used to guide the unsupervised learning. And due to the non-convex nature of the optimization problem for shapelet learning, K-means is used to find initial shapelet candidates. Different from existing shapelet/feature/dictionary learning algorithms, our method employs t-stochastic neighborhood embedding to overcome the crowding problem in projected low-dimensional space for shapelet learning. Moreover, our method can find an optimal length of the shapelets using $L_1$-norm regularization. The proposed method is then evaluated on the UCR time series dataset and an electrical component manipulation task, such as switching on, to prove its usefulness on improving clustering accuracy compared to other state-of-art feature learning algorithms in the robotic context.      
\end{abstract}

\/*
*/

\section{Introduction}
In a contact-rich manipulation task, such as grasping, assembly, and handling, different errors may occur. Interpreting and understanding the touch sensor information can help us identify the root causes. For instance, a force impulse is detected in a time interval that is not supposed to be present. This may indicate a possible collision of a robot on an obstacle in an environment. Another example would be a time delay of a certain event represented by a force pattern, such as a level shift. This could suggest a change in the environment to interact with. 

Uncovering and identifying certain force patterns can not only help us in diagnosing a robot application but also can be used for robot motion segmentation or detection of a schematic change in a robot application offline.  
  
\/*
\begin{figure}
    \centering
    \begin{minipage}[t]{0.1\textwidth}
        \raggedright
        \includegraphics[width=1.5\columnwidth]{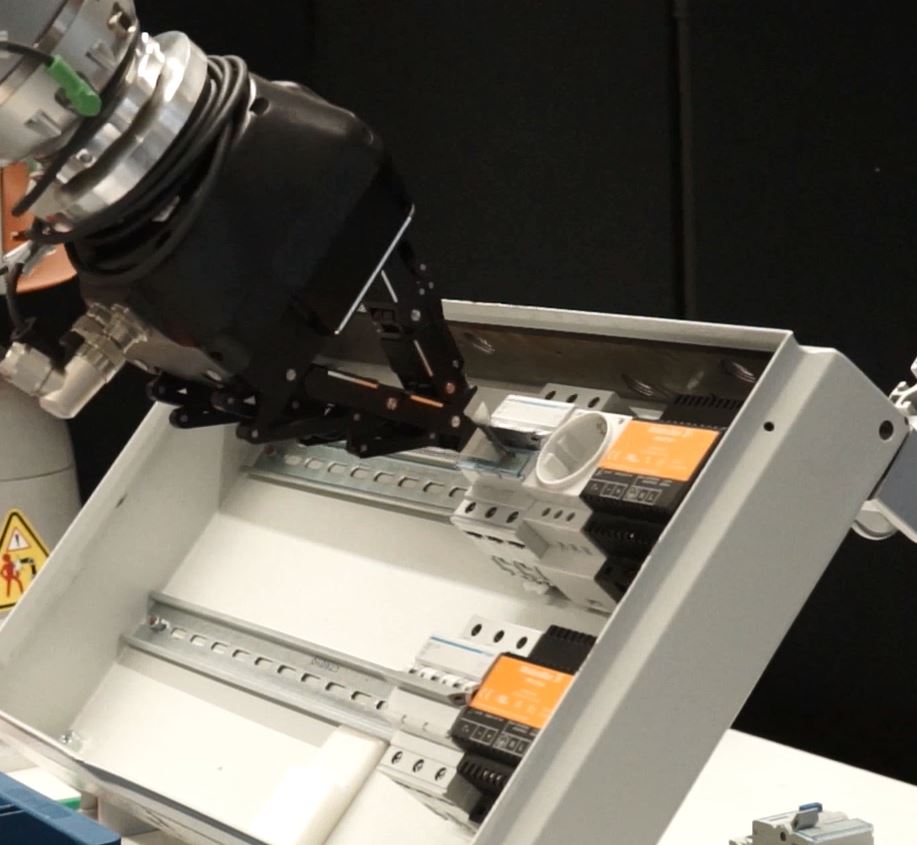} 
    \end{minipage}
    \begin{minipage}[t]{0.3\textwidth}
        
        \raggedleft
        \includegraphics[width=0.8\columnwidth]{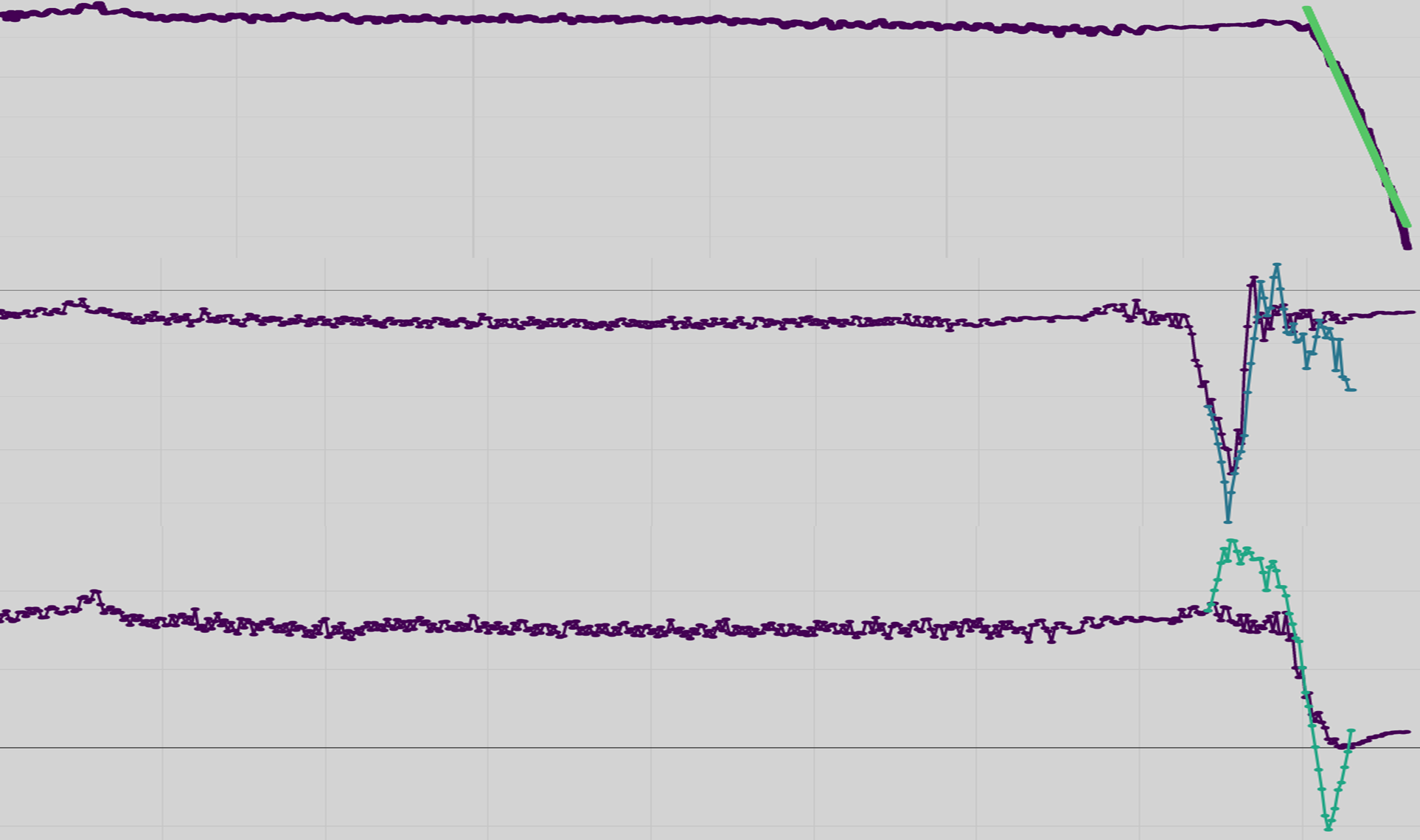} 
    \end{minipage}
    \caption{A robot application: switch pushing task (left) and examples of identifying discriminative time series sub-sequences that distinguish a time-series from others recorded in this application (right) 
    }
     \label{fig:intro1-robot_app}   
\end{figure}
*/

\begin{figure}%
    \centering
    \subfloat{{\includegraphics[width=2.5cm]{pics/SwitchOn.jpg} }}%
    \qquad
    \subfloat{{\includegraphics[width=5cm]{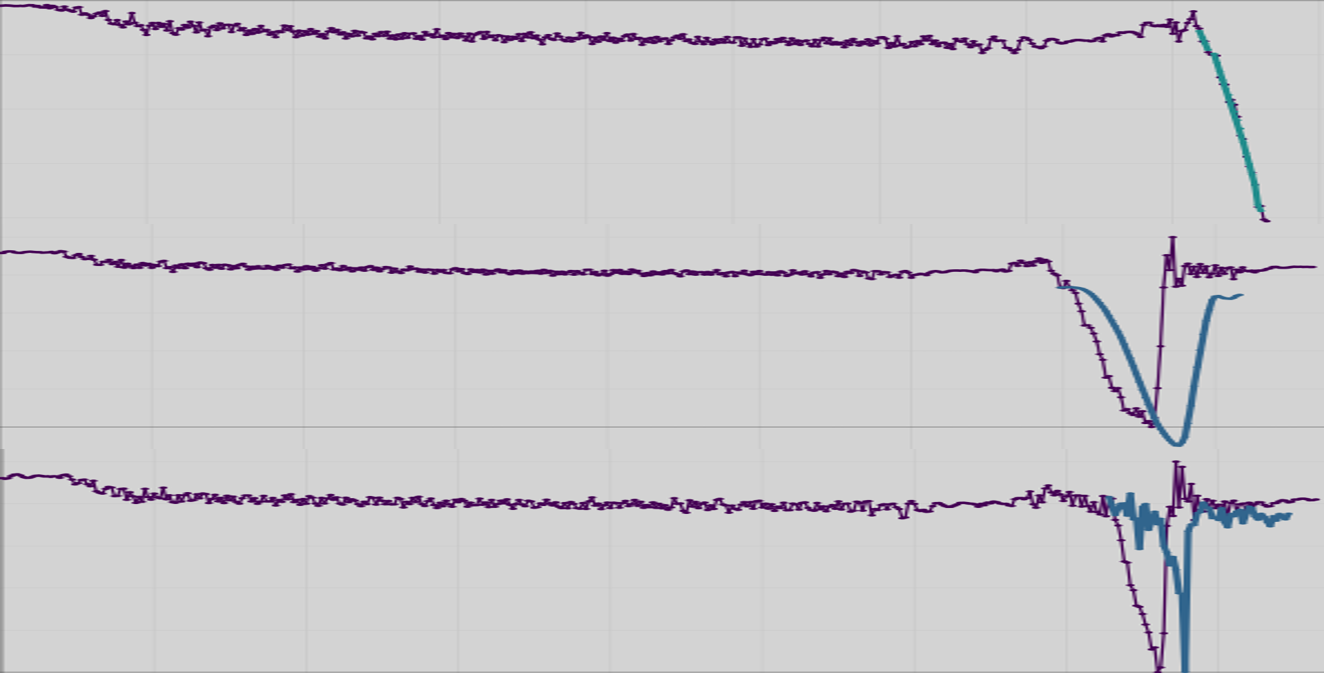} }}%
    \caption{A robot application: switch pushing task (left) and examples of identifying discriminative time series sub-sequences that distinguish a time series from others recorded in this application (right) }%
    \label{fig:intro1-robot_app}  
\end{figure}
\/*
\begin{figure}
    \centering
    \begin{minipage}[t]{0.1\textwidth}
        \includegraphics[width=1.5\columnwidth]{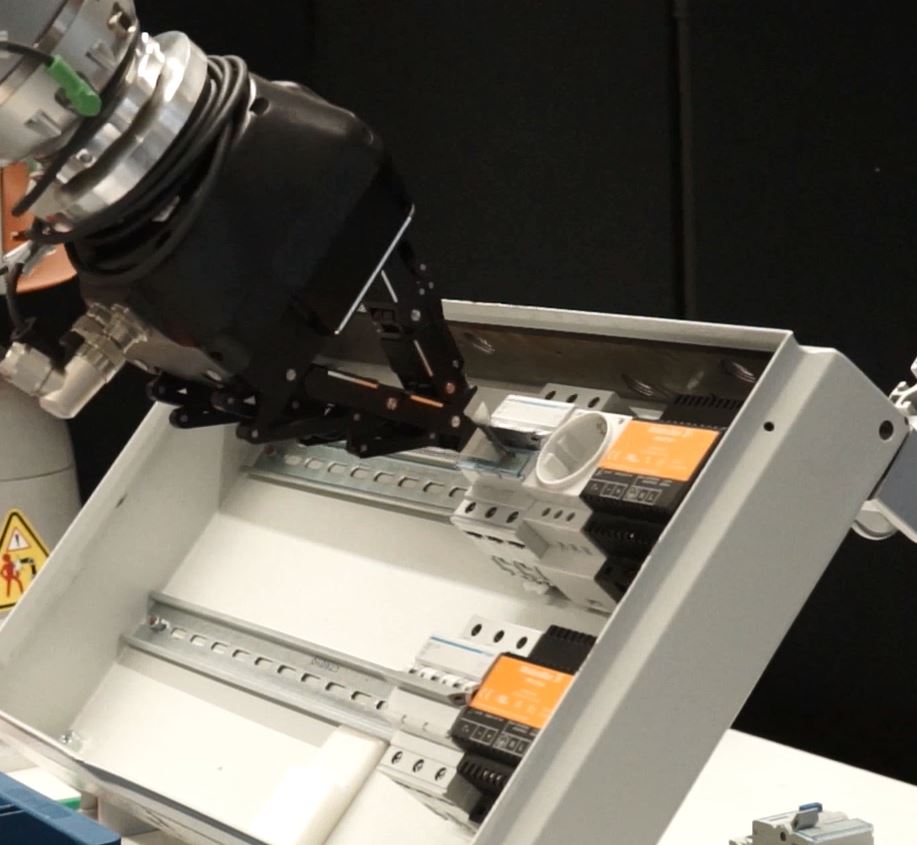} 
    \end{minipage}
    \begin{minipage}[t]{0.4\textwidth}
        
        \raggedleft
        \includegraphics[width=0.8\columnwidth]{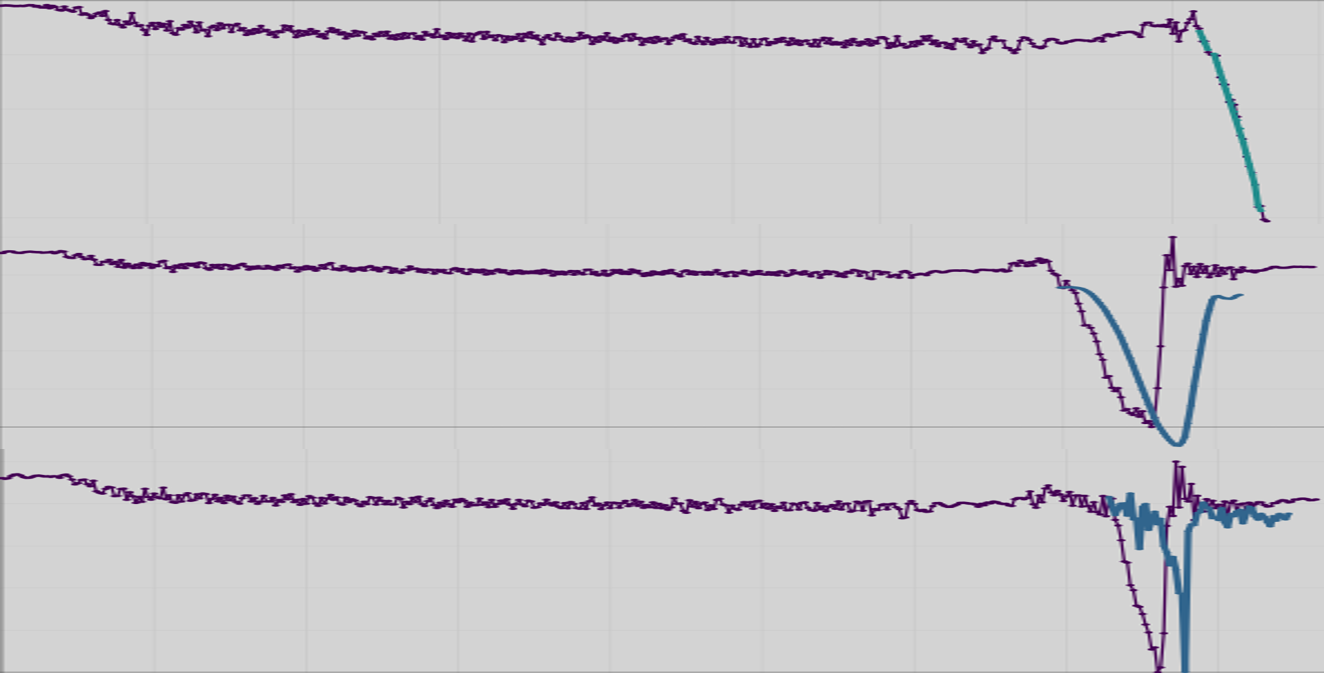} 
    \end{minipage}
    \caption{A robot application: switch pushing task (left) and examples of identifying discriminative time series sub-sequences that distinguish a time-series from others recorded in this application (right) 
    }
     \label{fig:intro1-robot_app}   
\end{figure}*/

An use case application is illustrated in Fig.\ref{fig:intro1-robot_app}, where a KUKA LBR iiwa 7kg tries to slide the lever up on the switch mounted on an electrical cabinet. Here two different errors can occur: the switch is already on, i.e., the lever of the switch is already pointing up; the switch is broken, where the lever can not be easily flipped on. The amplitude of external forces at the robot flange are depicted in Fig.\ref{fig:intro1-robot_app}, where the purple curves are the min-max normalized trajectories of external forces at robot flange from three different classes, with two classes for unknown error and one for a normal case.  The three additional plotted sub-sequences on the purple curves are learned shapelets, which highlight the differences between the trajectories from three different classes. In general, the extraction of time series sub-sequences, i.e., shapelets, can be termed as an unsupervised feature learning problem. Therefore, this paper aims at extracting discriminative features from the time-series data that can increase clustering accuracy. 

In summary, the contribution of this paper is stated as follows:
\begin{itemize}
\item This paper presents a novel approach for unsupervised shapelet learning in bag-of-words sense. 
\item In essence, this approach (NN-STNE) is to learn interpretable features in an unsupervised way. It outputs similarity measures between an original time series and a list of shapelets. Using these transformed similarity measures as input features for clustering UCR open time-series dataset, we prove that this feature learning algorithm achieves competitive results compared to other state-of-art feature selection algorithms.
\end{itemize}

\section{Related Work}\label{sec:related-work}


\textbf{Shapelets}\cite{Ye:2009:TSS:1557019.1557122} are originally proposed as time series sub-sequences that are considered as discriminating patterns for classification of temporal sequences.  The basic idea to discover shapelets is to assess all possible segments from time-series data based on a merit function that measures the predictive power of a given sub-sequence for some class labels. In \cite{Ye:2009:TSS:1557019.1557122} at first a large set of shapelet candidates are generated and their similarity to time series segments across all the training samples is computed using the brute-force algorithm. Then a decision tree algorithm is applied to recursively split training samples into different subsets by selecting a shapelet candidate to maximize the information gain for classification.
\/*
\textbf{Shapelets learning by clustering}
Instead of finding the optimal shapelets based on labeled data, the learning of shapelets can be considered as a process of dimension reduction or data compression with each shapelet representing a possible dimension or coordinate of time series. 
One of the best performing algorithm is K-Shape \cite{Paparrizos:2016:KEA:2949741.2949758}, which can learn shapelets that are shift and scale-invariant.
It is based on K-Means but uses normalized cross-correlation as a metric to measure time-series similarity that is shift-invariant and solves the maximization of Rayleigh Quotient for cluster centroid update. 
Recently, the artificial neural network is emerging due to its high representation power and its ability to learn complex cluster separation boundaries.
Inspired by \cite{Grabocka:2014:LTS:2623330.2623613} to use a neural network to learn shapelets, \cite{Zhang:2016:UFL:3060832.3060946} proposes to learn shapelets in an unsupervised way. Instead of using labeled data, \cite{Zhang:2016:UFL:3060832.3060946} enforces that similar input data should have similar pseudo-class labels by spectral analysis. It jointly optimizes candidate shapelets, class boundaries, and pseudo labels. 
*/
\textbf{Unsupervised feature selection} entails shapelet learning, because a shapelet can also be considered as a time series feature. The challenge of feature selection without class labels is to discover uncorrelated and discriminative features. Different algorithms have been introduced. One of the state-of-art algorithms is using dictionary learning to discover shapelet in a generative way \cite{DBLP:conf/case/ZhangLGWL19}. It considers a shapelet as an atom of dictionary and requires sliding the original time series into sub-sequences of the same length as shapelet, i.e., a dictionary atom. By trying to reconstruct slid sub-sequences, the algorithm jointly optimizes the shapelet dictionary and its corresponding sparse encoding for slid sub-sequences. This can result in losing global information of the original time-series and falling into the pitfalls of time series sub-sequence clustering \cite{Keogh03clusteringof}. Besides focusing on the shape feature of time series, each sample point in a time series can also be referred to as a potential feature similar to using pixels as input features for image clustering. \cite{Yang2011l21R} presents a $l_{2,1}$-norm regularized discriminative feature selection  algorithm for unsupervised learning (UDFS) . 

The presented algorithm in the scope of this paper differs from the above mentioned methods, in which we employ t-stochastic neighborhood embedding to perform a non-linear mapping between original data and shapelets and also attains the interpretability of the selected features.
 
\section{Preliminaries}
In this paper, scalar variables are denoted by unbold alphabets such as $(a,b,c, \alpha, \beta, \gamma, \cdots)$ whereas vector variables are denoted by bold lower-case alphabets $(\bm{a}, \bm{b}, \bm{c}, \bm{\alpha}, \cdots)$ and matrices by upper-case alphabets $(\bm{A}, \bm{B}, \cdots )$. 
The index of samples is represented by $n$; the index of a point in a time-series by $q$; the index of a point in a shapelet by $m$; the index of shapelets by $k$; the index for the class labels by $c$. Note that the unspecified parameters are denoted by lower-case letters while a constant defined number is represented with an upper-case letter, such as $N, C, K$, etc. For instance, the number of samples is denoted by $N$, the length of a time-series is written as $Q$, the length of patterns is represented with $M$, the number of patterns is $K$ and the total number of class labels is denoted by $C$.

\section{Unsupervised Shapelet Learning with Stochastic Neighborhood Embedding (SNE)}

\begin{figure}[htbp]
\centerline{\includegraphics[width=0.5\textwidth]{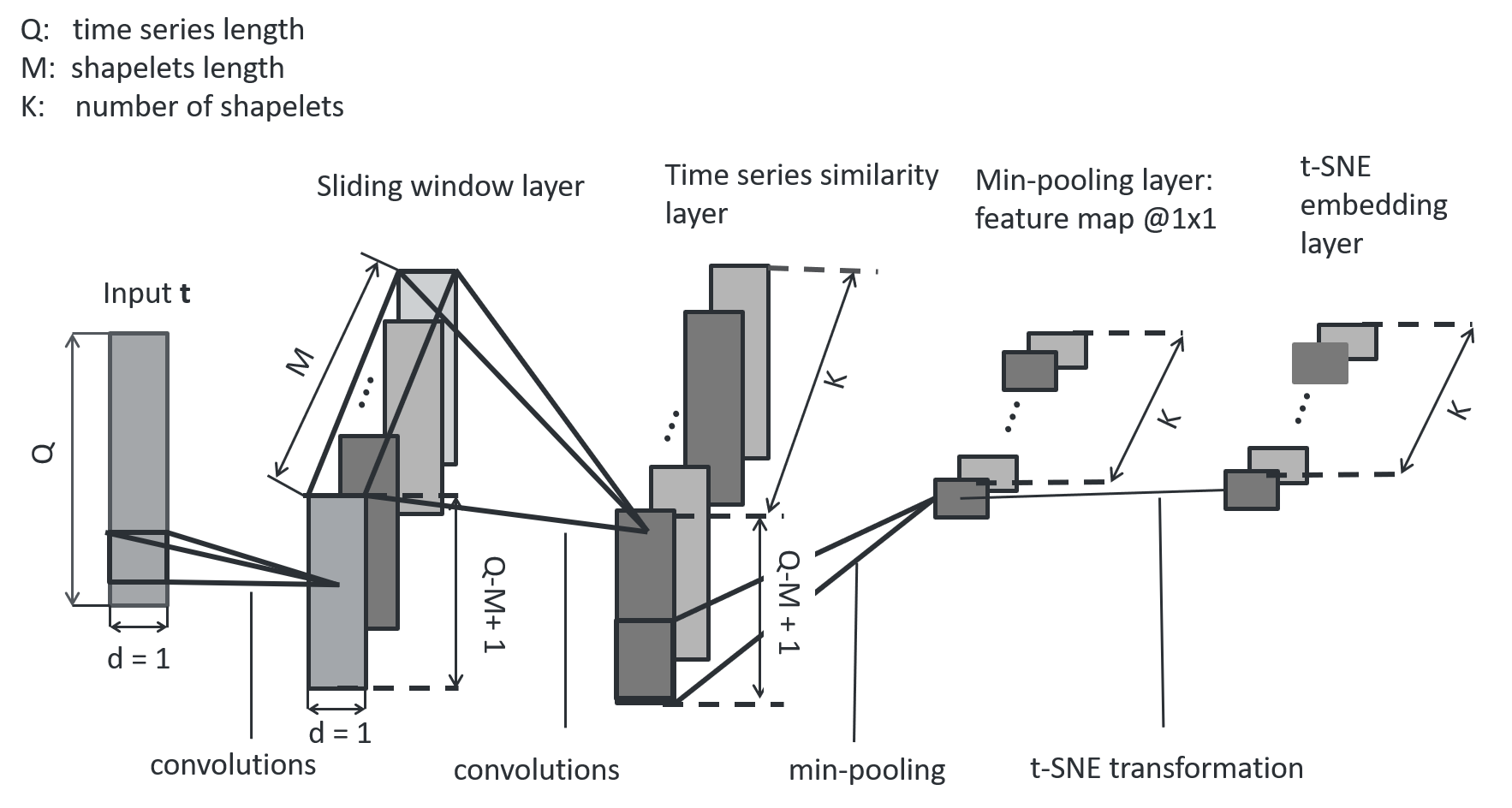}}
\caption{Overview of the network architecture}
\label{fig:fig_nn_overview}
\end{figure}

In this section, we aim to introduce the novel unsupervised shapelet learning model with deep embedding. Since a shapelet is a discriminative time-series segment, its length is always shorter than the input time series. Therefore, a sliding window approach from  \cite{Grabocka:2014:LTS:2623330.2623613}, which slides the input time series into equal size of shapelet length, is presented at first. Then follows the calculation of similarity between the candidate shapelets and the time series sub-sequences. After that, for each candidate shapelet, its corresponding most similar time series sub-subsequence from one input time series is selected. The normalized cross-correlation score for measuring the similarity between training data sub-subsequence and candidate shapelet is then converted into a probability distribution. The more similar they are, the higher the probability to assign the time series sub-sequence to this candidate shapelet will be.  
Based on this probability distribution, the corresponding shapelet candidates will be updated to an extent that is proportional to its assignment probability. Fig.\ref{fig:fig_nn_overview} illustrates the whole architecture at a glance.

\subsection{Layers in NN-STNE Network}
\textbf{Initial Estimation of Number and Length of Shapelets}

For the length of shapelets we adopt the same strategy as in \cite{Grabocka:2014:LTS:2623330.2623613}. By human inspection, we estimate a possible length of shapelets, e.g., $M$.  The number of shapelets, i.e., $K$ , of size $M$, is chosen in a way, such that a large number of input time series, i.e., $N$, of length $Q$ from different classes $C$ can be represented using only $K$ number of shapelets. It can be interpreted as a vector quantification problem with each shapelet as one bit. Then it is equivalent to solve the following equation:

\begin{equation}
2^{K } = N \times (Q - M) \times C \label{eq:eq_n_shp}
\end{equation}

By solving the above equation, one can obtain $K = \log_2 [N \times (Q - M) \times C] $ assuming that each shapelet is completely different from each other. 
 
\textbf{Sliding Window Approach}
Given a one-dimensional time-series sample, i.e. $\bm{t} = [t_1, t_2, t_3, \cdots, t_q]$ of length $Q$, and a desired shapelet to learn as $\bm{s} = [s_1,s_2, \cdots, s_m]$ of length $M$. 
In order to perform a point-to-point similarity measurement calculation, it is required that the segments to be compared should be of equal size. Since the expected length of a shapelet is often pre-defined according to some heuristics, the input time-series can be split into segments of equal length as shapelets. To achieve this,  the implementation of  \cite{Grabocka:2014:LTS:2623330.2623613} is used. The sliding window approach is realized by the convolution operation.
And therefore, a time series of length $Q$ can be divided into $J := Q - M +1$ sub-sequences of equal size of $M$. 

\textbf{Time Series Similarity Layer}
After sliding the input time-series into equal size, each will then be compared to a shapelet candidate.  There are many different metrics for measuring time-series similarity: cross-correlation \cite{Paparrizos:2016:KEA:2949741.2949758}, dynamic time warping \cite{salvador2004fastdtw}, Euclidean distance. In the scope of this paper, we adopt the normalized cross-correlation for measuring time series similarity. In this way, the obtained shapelets are not simply the average of matching time series sub-sequences from the input, but rather only a segment in a shapelet that better correlates with another segment in a slid sub-sequence from input time series get updated.  Consequently,  the characteristics in the input time series, such as, sharp changes, can be captured better. 
The calculation of cross normalized correlation in \cite{Paparrizos:2016:KEA:2949741.2949758} is applied and we use fast Fourier transform (FFT) to speed up the computation \cite{articleLewisFFT}.
By \cite{Paparrizos:2016:KEA:2949741.2949758}, the normalized cross correlation can be expressed as: \begin{equation}
\label{eq:ncc-4}
NCC(\bm{s}_k, \bm{t}_{i,j}) =   \underset{\omega}{\text{max}} \left( CC_\omega(\bm{s}^z_k, \bm{t}^z_{i,j})\right)
\end{equation} 
\begin{equation}
\label{eq:ncc-5}
D_{i,j,k}= 1 - NCC(\bm{s}_k, \bm{t}_{i,j})
\end{equation}
with $\bm{s}^z_k$, $\bm{t}^z_{i,j}$ as z-normalized $k$-th shapelet of length $M$ and $j$-th slid window of length $M$ from the $i$-th time series sample respectively. And $\omega$ is the amount of right or left shift of a time series sub-sequence according to the definition of cross correlation, which is expressed as a function $CC_{\omega}(\cdot)$. Additionally, $NCC(\bm{s}_k, \bm{t}_{i,j})$ is a function that denotes the normalized cross correlation between $k$-th shapelet $\bm{s}_k$  and time series sub-sequence $\bm{t}_{i,j}$ and $D_{i,j,k}$ is an entry in the time series similarity matrix $\bm{D}\in \mathbb{R}^{N\times J \times K}$. By the definition of normalized cross correlation, it has a range between -1 and 1, i.e., $NCC(\bm{s}_k, \bm{t}_{i,j}) \in \left[-1,1\right]$. For simplicity and avoid negative activation in neural network, we subtract the normalized cross correlation score from 1 to obtain $D_{i,j,k}$ as defined in (\ref{eq:ncc-5}).  The smaller the $D_{i,j,k}$, the more similar the sub-sequence and a shapelet candidate will be. 

\textbf{ T-Distributed Stochastic Neighborhood Embedding (t-SNE) Layer }

After transforming the time series data, i.e., $\bm{T} \in \mathbb{R}^{N\times Q}$ into distances to different shapelets, i.e., $\bm{D} \in \mathbb{R}^{N \times J \times K}$ 
Then follows the selection of for each shapelet  most matching sub-sequence in one time series sample by min-pooling, i.e., $F_{i,k} = \min_j \bm{D}$  with $D_{i,j,k}$ as an entry in $\bm{D}$ and  $F_{i,k}$ as an entry from the new matrix, i.e., $\bm{F} \in \mathbb{R}^{N \times K}$.
Therefore, we can represent a time series sample, i.e., $\bm{t}_i \in \mathbb{R}^Q$ by the distances to different shapelets, i.e., $\bm{f} \in \mathbb{R}^K$ as $i$-th row in $\bm{F} \in \mathbb{R}^{N \times K}$, where we reduce the information amount from original $Q$, i.e. the length of time-series, into $K$, i.e. the number of shapelets. Here a shapelet can be considered as a coordinate axis in lower-dimensional map space. 

When a high-dimensional data is mapped into a low-dimensional space, a crowding problem could occur \cite{DBLP:conf/nips/HintonR02}. 
Inspired by \cite{vandermaaten2008visualizing}, a t-student distribution is employed to convert the similarity between a most matching sub-sequence and  the shapelet candidates into probability: 
\begin{equation}
\label{eq:eq-tsne-2}
q_{i, k} = \frac{(1+ F_{i,k}) /\alpha)^{-\frac{\alpha + 1}{2}}}{\sum_k \left(1 + F_{i,k} /\alpha)^{-\frac{\alpha + 1}{2}}  \right)}
\end{equation}
where $\alpha $ is the degree of freedom of student $t$-distribution and $F_{i,k}$ is a distance metric always larger than $0$, and the smaller the distance metric, the higher the similarity score. In the following experiment, we let $\alpha = 1$ \cite{articleVanDerMaaten2009}.

\subsection{Objective Function}
\label{sec:objective_func}

\textbf{Spectral Analysis}. It is assumed that two similar time series should share similar distances to candidate shapelets.
To describe this, a mean square error scaled by Gaussian kernel from the spectral analysis is adopted \cite{Zhang:2016:UFL:3060832.3060946}.
Consider that $\bm{G} \in \mathbb{R}^{N \times N}$ is the matrix for describing similarity among all the time series samples with an entry defined as: $G_{\left(ij\right)} = \me^{-\frac{\norm{\bm{t}_i - \bm{t}_j}^2}{\sigma^2}}$ with $\sigma$ denotes the variance of the Gaussian kernel and $\bm{t}_i$, $\bm{t}_j$ are for two different time series samples. The variance of the Gaussian kernel also defines the effective number of neighbors for a given sample point\cite{NIPS2004_2619}. 
With this, the Gaussian kernel based mean square error is expressed as: 

\begin{equation}
\label{eq:eq-objective-2}
\begin{split}
&\frac{1}{2} \sum_{i = 1}^N \sum_{j = 1}^N G_{(ij)} \norm{\bm{q}_{(i,:)} - \bm{q}_{(j,:)}}_2^2 \\
&= \frac{1}{2} \sum_{k = 1}^K \sum_{i = 1}^N \sum_{j = 1}^N G_{(ij)} \left[q_{(k,i)} - q_{(k,j)}\right]^2 \\
&= \sum_{k = 1}^K \bm{q}_{(k,:)}^T \left( \bm{D}_G - G \right)\bm{q}_{(k,:)}\\
&= tr\left(\bm{q}^T\bm{L}_G\bm{q} \right)
\end{split}
\end{equation}
where $\bm{q}_{(i,:)},\bm{q}_{(j,:)}  \in \mathbb{R}^{1 \times K}$ are the transformed distances of time series sample $i, j$ to shapelet candidates respectively and $N$  is the number of time series samples. In addition, $\bm{L}_G = \bm{D}_G - \bm{G}$ is the Laplacian matrix in spectral analysis with $\bm{D}_G$ as a diagonal matrix with element defined as $D_G(i,i) = \sum_{j = 1}^n G_{(ij)}$
Besides this, it is also important that different shapelets should be as distinct to each other as possible. 

\textbf{Encouraging diverse shapelets}. Again we employ Gaussian kernel to penalize similar shapelets \cite{Zhang:2016:UFL:3060832.3060946}. The similarity between shapelets can be described as $\bm{H} \in \mathbb{R}^{K \times K}$, where an entry is defined as $H_{(i,j)} = \me^{-\frac{||\bm{s}_i - \bm{s}_j||^2}{\sigma^2}}$. And hence, to encourage distinct shapelet is to minimize the norm of shapelet similarity matrix, i.e., $||\bm{H}||_2^2$.

\textbf{Automatic selection of shapelets length}. Shapelets are the convolutional kernel weights in the time series similarity layer as shown in Fig.\ref{fig:fig_nn_overview}. To find out the optimal length of shapelet, it is equivalent to applying regularization techniques on kernel weights, where a $L_1$ norm is introduced to force shapelet filters to become zeros if possible. As a result, the zeros in shapelets will not contribute to the calculation of normalized cross-correlation. Consequently, the true length of shapelets can be obtained by removing zero values in the shapelet values.  

In summary, the total objective function to  minimize is formulated as:

\begin{equation}
\label{eq:eq-objective-4}
\mathcal{L} = tr\left(\bm{q}^T\bm{L}_G\bm{q} \right) + \lambda ||\bm{H}||_2^2 + \beta \sum_{k=1}^K \sum_{l=1}^M |s_{k,l}|
\end{equation}
with $\lambda$ as a weighting factor of the minimization of shapelet similarity and the last term $ \beta \sum_{k=1}^K \sum_{l=1}^M |s_{k,l}|$ as a shapelet regularization term to automatic select optimal  length of shapelets.  $\beta$ is a parameter given by the user to weigh the trade-off between learning of more similar shapelets to the input data and generalization. 
    
\section{Experiment and Evaluation}
\label{sec:experiment_and_evaluation}
In this section, the proposed method unsupervised shapelet learning with t-distributed stochastic neighborhood embedding layer is tested on public time-series dataset and its performance is evaluated compared to other unsupervised feature and shapelet learning techniques as mentioned in Sec. \ref{sec:related-work} such as 
Uncorrelated and Discriminative Feature Selection (UDFS)\cite{Yang2011l21R}, k-Shape \cite{Paparrizos:2016:KEA:2949741.2949758} and unsupervised shapelet learning \cite{Zhang:2016:UFL:3060832.3060946}.


\subsection{Data Sets}
To make our evaluation comparable to other state-of-art shapelet learning algorithms, a subset from the public open data sets UCR \footnote{http://timeseriesclassification.com/dataset.php} is used. On the other hand, we also need to prove its usefulness in the robotics application. Therefore, the data from the KUKA LBR iiwa switching on application as depicted in Fig.\ref{fig:intro1-robot_app} is used.  And a description of the used data-sets is presented in Table \ref{table:ucr-dataset}

\begin{table}[t]
\caption{Statistics of benchmark time series data set.}
\label{table:ucr-dataset}
\vskip 0.15in
\begin{center}
\begin{small}
\begin{sc}
\begin{tabular}{lcccr}
\toprule
Data set & Train/Test & Length & $\#$ Classes \\
\midrule
ECG 200 &  100/100 (200)&  96& 2\\
CBF    & 30/900 (930) & 128& 3 \\
Face Four    & 24/88 (112)& 350 &  4       \\
OSU Leaf    & 200/242(442)& 427 &        6 \\
\thead{Robot Switch \\ Push up} & 489/81(570) &433 & 3 \\
\bottomrule
\end{tabular}
\end{sc}
\end{small}
\end{center}
\vskip -0.1in
\end{table} 

\subsection{Evaluation Metrics}
To evaluate the performance of clustering, different evaluation metrics, such as Accuracy(ACC), Normalized Mutual Information(NMI)\cite{stgh02b}, can be applied. To have a comparative study on the method proposed in \cite{Zhang:2016:UFL:3060832.3060946}, we also use the Rand Index to evaluate our algorithm. 
\/*
The Rand Index (RI) defined as: 
\begin{equation}
\text{RI}=  \frac{\text{TP} + \text{TN}}{\text{TP} + \text{TN} + \text{FP} + \text{FN} }  = \frac{\text{TP}+\text{TN}}{ \binom{N}{2}}
\label{eq:eval_metrics}
\end{equation}
where TP and TN denote true positive and true negative respectively. A true positive is the number of times a pair of elements is grouped as in the ground truth. A true negative refers to the number of times a pair of elements is not grouped as in the ground truth. FP is false positive, which means the number of times a pair should not be assigned together but is clustered together by an algorithm, whereas FN is false negative, which refers to the number of times a pair should be assigned together based on the ground truth but is not grouped by a clustering algorithm. As denominator in Eq.\ref{eq:eval_metrics} is the total number of possible combinations of a pair of elements. */

\subsection{Comparison Results}
Since our algorithm transforms time series data into distances to shapelets and does not directly output pseudo-labels, to prove its usefulness, we consider the transformed shapelet distances as extracted features and feed them to a clustering method, such as KMeans. Consequently, a predicted label can be obtained. Then we compare our feature learning algorithms NN-TSNE with other feature learning algorithms mentioned in Sec.\ref{sec:related-work} such as, UDFS\cite{articleLi2020}, 
using Rand Index defined in  \cite{Zhang:2016:UFL:3060832.3060946}. We select the best results obtained by UDFS using different number of neighborhood and list the results in Table \ref{table:eval-res}. The best result for each data set is highlighted in bold.

\/*
\begin{table}[t]
\caption{Comparison of different algorithms in terms of clustering performance}
\label{table:eval-res}
\vskip 0.15in
\begin{center}
\begin{small}
\begin{sc}
\begin{tabular}{lcccr}
\toprule
Data set & KMeans  & \thead{ UDFS \\ +KMeans}&  \thead{ NN-STNE \\ +KMeans} \\
\midrule
ECG 200 &  0.6 &  0.55 &  0.7\\
CBF    & 0.74 & 0.73&0.93 \\
Face Four    & 0.74& 0.73 &  0.79 & 0.81       \\
OSU Leaf    & 0.76 & 0.76 &     0.75 \\
Average & x &x &x &x \\
\bottomrule
\end{tabular}
\end{sc}
\end{small}
\end{center}
\vskip -0.1in
\end{table} 
*/
\begin{table}[t]
\caption{Comparison of different algorithms in terms of clustering performance}
\label{table:eval-res}
\vskip 0.15in
\begin{center}
\begin{small}
\begin{sc}
\begin{tabular}{lcccr}
\toprule
Data set & KMeans  & \thead{ UDFS \\ +KMeans}  &  \thead{ NN-STNE \\ +KMeans} \\ 
\midrule
ECG 200 &  0.6 &  0.55   &\textbf{0.7}\\
CBF    & 0.74 & 0.73 & \textbf{0.93} \\
Face Four    & 0.74& 0.73   &\textbf{0.81}       \\  
OSU Leaf    & \textbf{0.76} & 0.76  &    0.76 \\
Switching Up & 0.74 &  0.74  & \textbf{1.0} \\
Average & 0.72 &  0.7  & \textbf{0.84} \\
\bottomrule
\end{tabular}
\end{sc}
\end{small}
\end{center}
\vskip -0.1in
\end{table} 
From Table \ref{table:eval-res} we can observe that in most of the cases using NN-STNE as feature selection algorithms before applying KMeans can help us achieve better clustering results than without applying any feature selection algorithms. Interesting is also to note that using UDFS as feature selection can somehow make the clustering result slightly worse. And from these five data sets, we can observe a $16.7\%$ of improvement on clustering results using KMeans on average.

\subsection{Conclusion}

Time series data can be analyzed from either the temporal perspective or shape perspective. In this paper, we focused on the shape perspective of time series data and proposed NN-STNE as a feature learning algorithm to discover discriminative time-series sub-sequences using embedded learning and proved that, when shape features in time series data prevail, applying it as a feature selection step can help improve clustering accuracy.
\/*

\subsection{Maintaining the Integrity of the Specifications}

The IEEEtran class file is used to format your paper and style the text. All margins, 
column widths, line spaces, and text fonts are prescribed; please do not 
alter them. You may note peculiarities. For example, the head margin
measures proportionately more than is customary. This measurement 
and others are deliberate, using specifications that anticipate your paper 
as one part of the entire proceedings, and not as an independent document. 
Please do not revise any of the current designations.

\section{Prepare Your Paper Before Styling}
Before you begin to format your paper, first write and save the content as a 
separate text file. Complete all content and organizational editing before 
formatting. Please note sections \ref{AA}--\ref{SCM} below for more information on 
proofreading, spelling and grammar.

Keep your text and graphic files separate until after the text has been 
formatted and styled. Do not number text heads---{\LaTeX} will do that 
for you.\cite{IEEEexample:articleetal}

\subsection{Abbreviations and Acronyms}\label{AA}
Define abbreviations and acronyms the first time they are used in the text, 
even after they have been defined in the abstract. Abbreviations such as 
IEEE, SI, MKS, CGS, ac, dc, and rms do not have to be defined. Do not use 
abbreviations in the title or heads unless they are unavoidable.

\subsection{Units}
\begin{itemize}
\item Use either SI (MKS) or CGS as primary units. (SI units are encouraged.) English units may be used as secondary units (in parentheses). An exception would be the use of English units as identifiers in trade, such as ``3.5-inch disk drive''.
\item Avoid combining SI and CGS units, such as current in amperes and magnetic field in oersteds. This often leads to confusion because equations do not balance dimensionally. If you must use mixed units, clearly state the units for each quantity that you use in an equation.
\item Do not mix complete spellings and abbreviations of units: ``Wb/m\textsuperscript{2}'' or ``webers per square meter'', not ``webers/m\textsuperscript{2}''. Spell out units when they appear in text: ``. . . a few henries'', not ``. . . a few H''.
\item Use a zero before decimal points: ``0.25'', not ``.25''. Use ``cm\textsuperscript{3}'', not ``cc''.)
\end{itemize}

\subsection{Equations}
Number equations consecutively. To make your 
equations more compact, you may use the solidus (~/~), the exp function, or 
appropriate exponents. Italicize Roman symbols for quantities and variables, 
but not Greek symbols. Use a long dash rather than a hyphen for a minus 
sign. Punctuate equations with commas or periods when they are part of a 
sentence, as in:
\begin{equation}
a+b=\gamma\label{eq}
\end{equation}

Be sure that the 
symbols in your equation have been defined before or immediately following 
the equation. Use ``\eqref{eq}'', not ``Eq.~\eqref{eq}'' or ``equation \eqref{eq}'', except at 
the beginning of a sentence: ``Equation \eqref{eq} is . . .''

\subsection{\LaTeX-Specific Advice}

Please use ``soft'' (e.g., \verb|\eqref{Eq}|) cross references instead
of ``hard'' references (e.g., \verb|(1)|). That will make it possible
to combine sections, add equations, or change the order of figures or
citations without having to go through the file line by line.

Please don't use the \verb|{eqnarray}| equation environment. Use
\verb|{align}| or \verb|{IEEEeqnarray}| instead. The \verb|{eqnarray}|
environment leaves unsightly spaces around relation symbols.

Please note that the \verb|{subequations}| environment in {\LaTeX}
will increment the main equation counter even when there are no
equation numbers displayed. If you forget that, you might write an
article in which the equation numbers skip from (17) to (20), causing
the copy editors to wonder if you've discovered a new method of
counting.

{\BibTeX} does not work by magic. It doesn't get the bibliographic
data from thin air but from .bib files. If you use {\BibTeX} to produce a
bibliography you must send the .bib files. 

{\LaTeX} can't read your mind. If you assign the same label to a
subsubsection and a table, you might find that Table I has been cross
referenced as Table IV-B3. 

{\LaTeX} does not have precognitive abilities. If you put a
\verb|\label| command before the command that updates the counter it's
supposed to be using, the label will pick up the last counter to be
cross referenced instead. In particular, a \verb|\label| command
should not go before the caption of a figure or a table.

Do not use \verb|\nonumber| inside the \verb|{array}| environment. It
will not stop equation numbers inside \verb|{array}| (there won't be
any anyway) and it might stop a wanted equation number in the
surrounding equation.

\subsection{Some Common Mistakes}\label{SCM}
\begin{itemize}
\item The word ``data'' is plural, not singular.
\item The subscript for the permeability of vacuum $\mu_{0}$, and other common scientific constants, is zero with subscript formatting, not a lowercase letter ``o''.
\item In American English, commas, semicolons, periods, question and exclamation marks are located within quotation marks only when a complete thought or name is cited, such as a title or full quotation. When quotation marks are used, instead of a bold or italic typeface, to highlight a word or phrase, punctuation should appear outside of the quotation marks. A parenthetical phrase or statement at the end of a sentence is punctuated outside of the closing parenthesis (like this). (A parenthetical sentence is punctuated within the parentheses.)
\item A graph within a graph is an ``inset'', not an ``insert''. The word alternatively is preferred to the word ``alternately'' (unless you really mean something that alternates).
\item Do not use the word ``essentially'' to mean ``approximately'' or ``effectively''.
\item In your paper title, if the words ``that uses'' can accurately replace the word ``using'', capitalize the ``u''; if not, keep using lower-cased.
\item Be aware of the different meanings of the homophones ``affect'' and ``effect'', ``complement'' and ``compliment'', ``discreet'' and ``discrete'', ``principal'' and ``principle''.
\item Do not confuse ``imply'' and ``infer''.
\item The prefix ``non'' is not a word; it should be joined to the word it modifies, usually without a hyphen.
\item There is no period after the ``et'' in the Latin abbreviation ``et al.''.
\item The abbreviation ``i.e.'' means ``that is'', and the abbreviation ``e.g.'' means ``for example''.
\end{itemize}
An excellent style manual for science writers is \cite{b7}.

\subsection{Authors and Affiliations}
\textbf{The class file is designed for, but not limited to, six authors.} A 
minimum of one author is required for all conference articles. Author names 
should be listed starting from left to right and then moving down to the 
next line. This is the author sequence that will be used in future citations 
and by indexing services. Names should not be listed in columns nor group by 
affiliation. Please keep your affiliations as succinct as possible (for 
example, do not differentiate among departments of the same organization).

\subsection{Identify the Headings}
Headings, or heads, are organizational devices that guide the reader through 
your paper. There are two types: component heads and text heads.

Component heads identify the different components of your paper and are not 
topically subordinate to each other. Examples include Acknowledgments and 
References and, for these, the correct style to use is ``Heading 5''. Use 
``figure caption'' for your Figure captions, and ``table head'' for your 
table title. Run-in heads, such as ``Abstract'', will require you to apply a 
style (in this case, italic) in addition to the style provided by the drop 
down menu to differentiate the head from the text.

Text heads organize the topics on a relational, hierarchical basis. For 
example, the paper title is the primary text head because all subsequent 
material relates and elaborates on this one topic. If there are two or more 
sub-topics, the next level head (uppercase Roman numerals) should be used 
and, conversely, if there are not at least two sub-topics, then no subheads 
should be introduced.

\subsection{Figures and Tables}
\paragraph{Positioning Figures and Tables} Place figures and tables at the top and 
bottom of columns. Avoid placing them in the middle of columns. Large 
figures and tables may span across both columns. Figure captions should be 
below the figures; table heads should appear above the tables. Insert 
figures and tables after they are cited in the text. Use the abbreviation 
``Fig.~\ref{fig}'', even at the beginning of a sentence.

\begin{table}[htbp]
\caption{Table Type Styles}
\begin{center}
\begin{tabular}{|c|c|c|c|}
\hline
\textbf{Table}&\multicolumn{3}{|c|}{\textbf{Table Column Head}} \\
\cline{2-4} 
\textbf{Head} & \textbf{\textit{Table column subhead}}& \textbf{\textit{Subhead}}& \textbf{\textit{Subhead}} \\
\hline
copy& More table copy$^{\mathrm{a}}$& &  \\
\hline
\multicolumn{4}{l}{$^{\mathrm{a}}$Sample of a Table footnote.}
\end{tabular}
\label{tab1}
\end{center}
\end{table}

\begin{figure}[htbp]
\caption{Example of a figure caption.}
\label{fig}
\end{figure}

Figure Labels: Use 8 point Times New Roman for Figure labels. Use words 
rather than symbols or abbreviations when writing Figure axis labels to 
avoid confusing the reader. As an example, write the quantity 
``Magnetization'', or ``Magnetization, M'', not just ``M''. If including 
units in the label, present them within parentheses. Do not label axes only 
with units. In the example, write ``Magnetization (A/m)'' or ``Magnetization 
\{A[m(1)]\}'', not just ``A/m''. Do not label axes with a ratio of 
quantities and units. For example, write ``Temperature (K)'', not 
``Temperature/K''.

\section*{Acknowledgment}

The preferred spelling of the word ``acknowledgment'' in America is without 
an ``e'' after the ``g''. Avoid the stilted expression ``one of us (R. B. 
G.) thanks $\ldots$''. Instead, try ``R. B. G. thanks$\ldots$''. Put sponsor 
acknowledgments in the unnumbered footnote on the first page.

\section*{References}

Please number citations consecutively within brackets \cite{b1}. The 
sentence punctuation follows the bracket \cite{b2}. Refer simply to the reference 
number, as in \cite{b3}---do not use ``Ref. \cite{b3}'' or ``reference \cite{b3}'' except at 
the beginning of a sentence: ``Reference \cite{b3} was the first $\ldots$''

Number footnotes separately in superscripts. Place the actual footnote at 
the bottom of the column in which it was cited. Do not put footnotes in the 
abstract or reference list. Use letters for table footnotes.

Unless there are six authors or more give all authors' names; do not use 
``et al.''. Papers that have not been published, even if they have been 
submitted for publication, should be cited as ``unpublished'' \cite{b4}. Papers 
that have been accepted for publication should be cited as ``in press'' \cite{b5}. 
Capitalize only the first word in a paper title, except for proper nouns and 
element symbols.

For papers published in translation journals, please give the English 
citation first, followed by the original foreign-language citation \cite{b6}.

*/
\bibliographystyle{IEEETran}
\bibliography{IEEEabrv,icra-references}

\/*
\vspace{12pt}
\color{red}
IEEE conference templates contain guidance text for composing and formatting conference papers. Please ensure that all template text is removed from your conference paper prior to submission to the conference. Failure to remove the template text from your paper may result in your paper not being published.
*/

\end{document}